\documentclass{article}

\usepackage[preprint]{corl_2026} 

\usepackage{amsmath}    
\usepackage{amssymb}    
\usepackage{graphicx}   
\usepackage{booktabs}   
\usepackage{multirow}   
\usepackage{xcolor}  
\usepackage{pifont}
\usepackage{marvosym}

\usepackage{booktabs}
\usepackage{multirow}
\usepackage[table]{xcolor}
\usepackage{adjustbox}
\usepackage{array}
\usepackage{siunitx}
\usepackage{threeparttable}
\usepackage{float}

\usepackage{graphicx}
\usepackage{booktabs}
\usepackage{longtable,booktabs,xcolor,colortbl}
\definecolor{headerblue}{RGB}{221,235,247}
\definecolor{lightgray}{RGB}{245,245,245}

\definecolor{HeaderGray}{HTML}{E9E9E9}
\definecolor{L1Shade}{HTML}{EEF7FF}    
\definecolor{L3Shade}{HTML}{FFF3EC}    

\newcolumntype{N}{S[table-format=3.1]}
\newcolumntype{B}{>{\columncolor{L1Shade}}S[table-format=3.1]}
\newcolumntype{P}{>{\columncolor{L3Shade}}S[table-format=3.1]}

\definecolor{scenepink}{RGB}{248,224,230}

\newcommand{\scenerow}[1]{%
\rowcolor{scenepink}
\multicolumn{2}{@{}p{\dimexpr\textwidth\relax}@{}}{%
\footnotesize\textbf{#1}} \\
}

\title{RoboFollow: Unveiling the Instruction Following Mirage in Embodied Agents}

\title{RoboFollow: Unveiling the Instruction Following Mirage in Embodied Agents}

\author{
\textbf{Chang Guo$^{1}$ \quad Yukun Xie$^{1}$ \quad Bohan Tan$^{1}$ \quad Zheng Chang$^{1}$ \quad Zhaokai Yin$^{1}$}\\
\textbf{Qianli Ma$^{1}$ \quad Yingqiao Wang$^{1}$ \quad Chao Liang$^{2}$ \quad Zhipeng Zhang$^{1,\dagger}$}\\
$^{1}$AutoLab, School of Artificial Intelligence, Shanghai Jiao Tong University\\
$^{2}$Research Lab, Anyverse Dynamics\\
\texttt{zhipengzhang@sjtu.edu.cn}
}

\begin{document}
\maketitle

\begingroup
\renewcommand{\thefootnote}{\fnsymbol{footnote}}
\footnotetext[2]{Corresponding Author.}
\endgroup

\begin{abstract}
Modern embodied agents achieve impressive success rates, yet their actual instruction-following ability is far weaker than these numbers suggest. We trace this illusion to a structural property we term low \emph{scene entropy}: when a visual scene admits only one valid task, language becomes redundant and a policy can score highly while barely using it. We introduce \textbf{RoboFollow}, a diagnostic benchmark with three principles: (\textbf{1}) \textbf{High Scene Entropy}: each training scene supports multiple kinematically distinct task branches, making vision alone insufficient and forcing reliance on language. (\textbf{2}) \textbf{Hierarchical Diagnostic Protocol}: a four-level protocol (L0--L3) progressively perturbs visual layout and semantics, probing whether equivalent instructions yield consistent behavior and distinct ones yield discriminable behavior across spatial relations, attributes, trajectory constraints, and logic. (\textbf{3}) \textbf{Confound-Controlled Diagnosis}: we simplify interaction objects, restrict actions to the trained repertoire and report stage-wise Intent and Execution scores, isolating comprehension from motor execution. Evaluation of nine VLA and WAM policies shows that strong L0 performance, where attained, does not reliably transfer to L1--L3 under our fine-tuning setup. Representative mitigations, including stronger VLM backbones, QA co-training, LangForce, and Classifier-Free Guidance, all fail to close this gap. RoboFollow exposes genuine instruction following as a critical, overlooked bottleneck. Code and dataset are available at https://github.com/AutoLab-SAI-SJTU/RoboFollow and https://huggingface.co/datasets/AutoLab-SJTU/robofollow-data.
\end{abstract}

\section{Introduction}
\label{sec:intro}

Language-conditioned robot policies are increasingly expected to serve as general-purpose embodied agents. Recent Vision-Language-Action (VLA)~\cite{zitkovich2023rt,kim2024openvla,black2024pi_0} and World Action Models (WAM)~\cite{wu2023unleashing,bi2025motus} have achieved strong success rates on manipulation benchmarks, while modular embodied systems increasingly use language as the interface between human intent and physical execution~\cite{feng2025multi}. However, high task success does not necessarily imply instruction following. A policy may complete a task because the visual scene already suggests a plausible action, while the language instruction is ignored, weakly used, or treated merely as a task identifier. For deployable robots, this distinction is critical: a visually successful action can still be semantically wrong.

This problem is difficult to expose with standard evaluation protocols. Most manipulation benchmarks emphasize final task success, which entangles visual recognition, language grounding, planning, and low-level control. More importantly, many benchmark episodes contain a dominant or canonical behavior that can be inferred from the initial visual observation, making language partially redundant. Such benchmarks remain valuable for measuring manipulation competence, but they are less diagnostic of whether a policy uses language to choose among multiple feasible behaviors. Recent studies further show that VLA policies can be insensitive to linguistic perturbations on existing benchmarks~\cite{xu2025seeing,zhou2025libero,fei2025liberoplusindepthrobustnessanalysis}. RoboFollow complements these evaluations by constructing task ambiguity during training and diagnosing grounding through controlled splits and stage-wise scoring.

To address this gap, we introduce \textbf{RoboFollow}, a diagnostic benchmark for evaluating whether embodied agents use language to select and execute the intended behavior. RoboFollow is built around a simple principle: language should be necessary rather than optional. It constructs high-ambiguity scenes in which the same or highly similar visual configuration supports multiple semantically valid and kinematically feasible task branches. In such scenes, vision alone is insufficient to identify the intended behavior, forcing the policy to rely on the instruction for disambiguation. We quantify this ambiguity through \emph{scene entropy}, the conditional entropy of training task labels given the task-independent initial scene specification. 

RoboFollow evaluates instruction following along three complementary dimensions. First, it covers four scene families: spatial relations, object attributes and action selection, trajectory and orientation constraints, and elementary logical grounding. Second, it introduces a four-level evaluation protocol: L0 measures in-distribution performance, L1 tests visual grounding under changed layouts, L2 tests semantic recombination under fixed layouts, and L3 combines visual and semantic perturbations. Third, it separates semantic misunderstanding from motor failure through stage-wise \emph{Intent} and \emph{Execution} Scores. Intent measures whether the policy selects the correct object, relation, waypoint, or logical branch, while Execution measures whether the selected subgoal is physically completed.

We conduct a systematic evaluation of representative VLA and WAM models on RoboFollow. Across models and scene families, we observe a consistent pattern: models achieve strong in-distribution performance at L0, but their Intent Scores degrade sharply under L1--L3. We further examine stronger VLM backbones, QA co-training, LangForce, and classifier-free guidance, but none closes the generalization gap beyond L0. These results suggest that robust language-conditioned task selection remains a challenge for the evaluated policies after fine-tuning.

Our contributions can be summarized as follows:
\begin{enumerate}
\setlength{\itemsep}{2pt}
\setlength{\parskip}{0pt}
\setlength{\parsep}{0pt}
\setlength{\topsep}{2pt}

\item \textbf{A language-necessary diagnostic benchmark with controlled execution confounds.}
We introduce RoboFollow, which constructs high-entropy scenes where language is required
to disambiguate among multiple feasible task branches. By using simple objects, short-horizon
interactions, and action primitives covered by training, RoboFollow reduces motor-execution
confounds and enables a targeted diagnosis of semantic instruction following.

\item \textbf{A hierarchical diagnostic protocol with stage-wise scoring.}
We design L0--L3 evaluation splits that progressively test in-distribution execution,
visual grounding under layout changes, semantic recombination under familiar visual contexts,
and joint visual-semantic generalization. We further report stage-wise Intent and Execution
Scores to separate errors in intent selection from execution inaccuracies conditioned on a
correct intent.

\item \textbf{A systematic analysis of current models and optimizations.}
We evaluate representative VLA and WA models, together with stronger VLM backbones,
QA co-training, LangForce, and classifier-free guidance, and show that these strategies
remain insufficient for robust instruction following.

\end{enumerate}

    \section{Related Work}
    
    \subsection{Vision-Language-Action and World Action Models}
    
    Vision-Language-Action (VLA) models~\cite{zitkovich2023rt,kim2024openvla,black2024pi_0,nvidia2025gr00tn1openfoundation,intelligence2025pi05visionlanguageactionmodelopenworld,zhang2026vlm4vlarevisitingvisionlanguagemodelsvisionlanguageaction} extend pre-trained VLMs to connect internet-scale perception with physical execution across diverse datasets~\cite{embodimentcollaboration2025openxembodimentroboticlearning}. Early methods discretize continuous actions for autoregressive generation~\cite{RT1,zitkovich2023rt,octomodelteam2024octoopensourcegeneralistrobot,kim2024openvla}, whereas recent architectures improve precision through flow matching~\cite{black2024pi_0,zheng2025xvlasoftpromptedtransformerscalable,nvidia2025gr00tn1openfoundation}, physically grounded representations~\cite{chen2025internvlam1spatiallyguidedvisionlanguageaction,qu2025spatialvlaexploringspatialrepresentations,zhang2025tavlaelucidatingdesignspace}, efficient quantization~\cite{shukor2025smolvlavisionlanguageactionmodelaffordable,xu2026qvlachannelsequalvisionlanguageaction}, and visual chain-of-thought reasoning~\cite{intelligence2025pi05visionlanguageactionmodelopenworld,zhao2025cotvlavisualchainofthoughtreasoning,zhang2026vlm4vlarevisitingvisionlanguagemodelsvisionlanguageaction}. 
    World Action Models (WAM) move beyond reactive visuomotor mapping by coupling action generation with predictive modeling of environment dynamics. This paradigm advances from diffusion-based visuomotor policies~\cite{chi2024diffusionpolicyvisuomotorpolicy,wang2024hierarchicaldiffusionpolicymanipulation,xue2025reactivediffusionpolicyslowfast} to predictive world models that jointly model visual dynamics and actions, using video generation to forecast future states and infer the corresponding actions~\cite{li2025unifiedvideoactionmodel,kim2026cosmospolicyfinetuningvideo,ye2026worldactionmodelszeroshot,chen2025largevideoplannerenables,bi2025motus,pai2025mimicvideovideoactionmodelsgeneralizable,li2026causalworldmodelingrobot}.
    
    \subsection{Benchmarks for Robotic Manipulation Evaluation}
    
    Robotic manipulation benchmarks span simulation and real-world evaluation. RLBench~\cite{james2019rlbenchrobotlearningbenchmark} and SimplerEnv~\cite{li2024evaluatingrealworldrobotmanipulation} provide standardized control protocols, while CALVIN~\cite{mees2022calvin}, VIMA~\cite{jiang2023vimageneralrobotmanipulation}, VLABench~\cite{zhang2024vlabenchlargescalebenchmarklanguageconditioned}, and RoboTwin2.0~\cite{chen2025robotwin20scalabledata} target long-horizon interaction, multimodal reasoning, semantic generalization, and scalable demonstration generation~\cite{zhang2024vlabenchlargescalebenchmarklanguageconditioned,mu2025robotwindualarmrobotbenchmark,chen2025robotwin20scalabledata,nasiriany2024robocasalargescalesimulationeveryday,yakefu2025robochallengelargescalerealrobotevaluation}. LIBERO~\cite{liu2023liberobenchmarkingknowledgetransfer}, LIBERO-PRO~\cite{zhou2025libero}, and LIBERO-Plus~\cite{fei2025liberoplusindepthrobustnessanalysis} further study knowledge transfer and robustness, showing that policies often exploit visual patterns instead of language semantics. RoboFollow explicitly pairs shared training scenes with multiple task specifications. This construction addresses training-time task ambiguity, complementing the test-time perturbations studied in LIBERO-PRO. $\pi_{0.7}$ reports stronger instruction following with more diverse data and richer contextual conditioning~\cite{intelligence2026pi07}; these different training and evaluation settings motivate our diagnostic. IVA addresses false-premise verification and correction~\cite{hsieh2025teachingiva}, complementing RoboFollow's selection among feasible tasks.
\section{RoboFollow Benchmark}
\label{sec:robofollow}

RoboFollow is designed to evaluate whether embodied agents use language to
select and execute the intended behavior, rather than relying on visual shortcuts. Its central design principle is to make language necessary for task
identification. In each diagnostic scene, the same or highly similar visual
configuration supports multiple semantically valid and kinematically feasible
task branches. Therefore, the initial observation alone is insufficient to infer the target behavior, and the policy must condition on the instruction to resolve task ambiguity.

Concretely, RoboFollow contains four diagnostic scene families comprising 75 training task labels grouped into six task-independent initial scene configurations. Let $T$ denote a training task label (shared by its paraphrases) and $S$ the task-independent initial scene specification, including objects, placement distributions, and observable state predicates, excluding instructions and goals. We define
\begin{equation}
H_{\mathrm{scene}}=H_{\mathrm{train}}(T\mid S)
=-\sum_s p(s)\sum_t p(t\mid s)\log_2p(t\mid s).
\label{eq:scene_entropy}
\end{equation}
Here, $p(s)$ is the fraction of training demonstrations associated
with scene specification $s$, and $p(t\mid s)$ is the fraction of
demonstrations within that scene group labeled with task $t$.

With equal demonstrations per task, $H_{\mathrm{scene}}=\sum_s(K_s/K)\log_2K_s$, where $K_s$ counts task labels sharing scene specification $s$ and $K=\sum_sK_s$. RoboFollow has $K_s=(16,16,16,16,7,4)$ and $K=75$, yielding $3.782$ bits, compared with $0.880$ bits for the equally weighted LIBERO Spatial/Object/Goal/Long suites under the same task-label definition. Grouping details are given in Appendix~\ref{sec:entropy_details}. We use scene entropy as a dataset design principle rather than as
a final evaluation metric: its role is to remove visual shortcuts during
training and evaluation. After this ambiguity is established, RoboFollow
diagnoses instruction following through controlled test splits and stage-wise
Intent and Execution scores.




\subsection{Scene Design}
\label{sec:scene_design}

\begin{figure*}[t]
  \centering
  \includegraphics[width=0.95\textwidth]{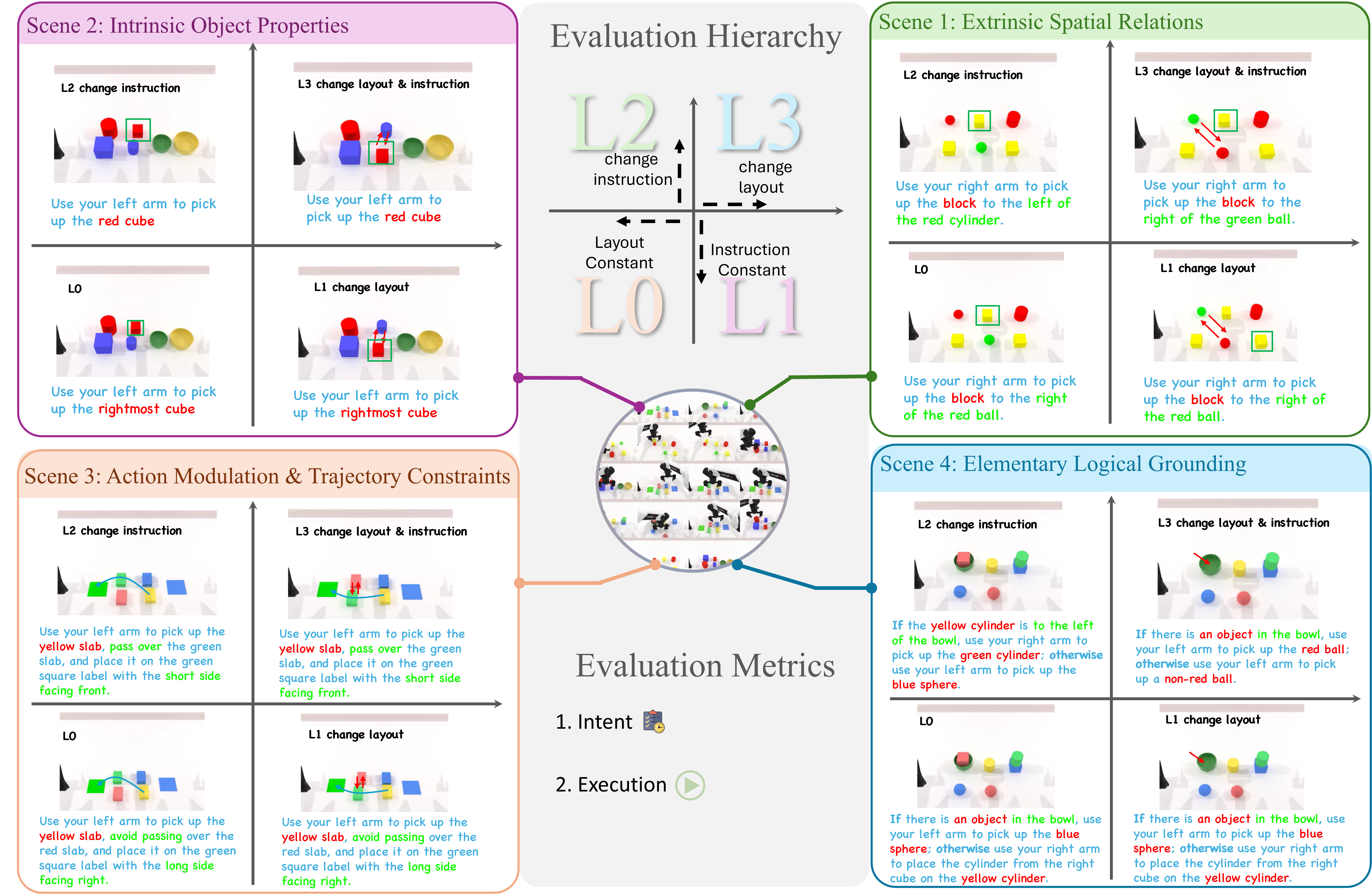}
  \caption{\textbf{Overview of RoboFollow scene design.}
  RoboFollow contains four diagnostic scenes targeting extrinsic spatial relations,
  intrinsic object properties, action and trajectory constraints, and elementary
  logical grounding. Each scene is evaluated under four levels, from in-distribution
  testing to combined visual-semantic perturbations.}
  \label{fig:scene_overview}
  \vspace{-0.8em}
\end{figure*}

RoboFollow evaluates models with four hierarchical test levels. These levels are
not intended as a strict difficulty ordering; instead, they isolate different
generalization dimensions. Overview of RoboFollow scene design is shown in Fig.~\ref{fig:scene_overview}, more details are provided in Appendix~\ref{sec:supp_benchmark_details}. 

\ding{170} \textbf{L0 (In Distribution):}
The test configuration follows the training distribution and measures standard
in-distribution performance.

\ding{170} \textbf{L1 (Visual Grounding):}
The instruction remains unchanged, but object positions are swapped. This tests
whether the model binds language to the correct physical entities rather than
memorizing spatial coordinates.

\ding{170} \textbf{L2 (Semantic Compositionality):}
The visual layout remains unchanged, but the instruction uses novel
recombinations of semantic attributes seen during training. This tests whether
the model captures atomic meanings rather than memorizing specific
text--object or attribute--object pairings.

\ding{170} \textbf{L3 (Visual-Semantic Mixture):}
Both the visual layout and the instruction are perturbed, evaluating whether the
model can jointly handle visual grounding and semantic recombination.

All training and validation splits are constructed to preserve linguistic clarity, avoid leakage across evaluation levels, and ensure that test-time target actions remain within the demonstrated behavioral repertoire. We next provide a concise overview of the four diagnostic scenes. 

\textbf{Scene 1: Extrinsic Spatial Relations.}
Scene 1 evaluates whether agents can ground extrinsic spatial relations such as ``to the left of'', ``to the right of'', and ``behind''. The core challenge is that multiple candidate objects are visually similar or identical, so the target cannot be determined from appearance alone. Instead, the policy must identify the correct object by binding relational language to the current spatial configuration. This scene therefore tests whether the model truly understands spatial prepositions, rather than memorizing absolute object positions or canonical layouts.


\textbf{Scene 2: Intrinsic Object Properties.}
Scene 2 evaluates compositional grounding over intrinsic object properties and action primitives. Instructions specify combinations of attributes such as color, size, and shape, together with actions such as pick, push, stack, and place. The object sets are designed so that no single attribute is always sufficient for identifying the target, requiring the model to compose multiple semantic cues. When necessary, kinematic choices such as arm selection are made explicit in the instruction, reducing ambiguity from physical feasibility.


\textbf{Scene 3: Fine-Grained Action Modulation.}
Scene 3 evaluates whether models can follow procedural constraints beyond achieving a final goal state. In this scene, instructions specify not only the source object and destination, but also intermediate motion constraints such as waypoints and final orientations. This design tests whether the policy follows the instructed procedure, rather than merely producing an action that ends in a plausible final configuration. It is particularly useful for diagnosing whether models treat language as a coarse task label or as a fine-grained control signal.



\textbf{Scene 4: Elementary Logical Grounding.}
Scene 4 evaluates elementary logical forms required for robust instruction following, including temporal sequencing, explicit negation, and observable conditional branching. Unlike long-horizon planning benchmarks, this scene focuses on short and controlled instructions such as ``first do A then do B'',
``not A'', and ``if A then do B else do C''. The relevant scene state is varied across splits, so the policy must evaluate the current observation rather than memorizing fixed instruction--trajectory pairs.


\subsection{Metrics}
\label{sec:metrics}

Binary task success based on the final environment state is often insufficient for diagnosing instruction following. It may penalize a policy that selects the correct intent but fails due to a minor execution error, while rewarding a policy that reaches the final state after semantically incorrect intermediate actions. RoboFollow therefore adopts a \textbf{Multi-Stage Intent-Execution Scoring} framework, where each task is decomposed into sequential sub-stages and evaluated using two complementary scores. More details are provided in Appendix~\ref{sec:supp_metrics_details}. 

\ding{170} \textbf{Intent Score (Semantic Grounding):}
Intent Score measures whether the policy selects the correct semantic target at each stage, such as the intended object, relation, waypoint, action primitive, orientation, or logical branch. It focuses on whether the policy attempts the behavior specified by the instruction, even when physical execution is imperfect.

\ding{170} \textbf{Execution Score (Kinematic Proficiency):}
Execution Score measures whether the corresponding subgoal is successfully completed, such as grasping, transporting, or placing the selected object.

By separating intent from execution, RoboFollow distinguishes semantic misunderstanding from low-level control failure and penalizes policies that achieve the final state through incorrect intermediate actions. A dedicated finishing stage accounts for 20\% of the score, requiring policies to refrain from further actions unrelated to the instruction after completing the requested task in order to receive full credit.

Completion Rate (CR) complements IS and ES with an action-dependent binary completion
criterion. Within each scene and difficulty level, CR is the unweighted mean of
per-task completion rates. It does not require every intermediate stage
to receive full credit. The exact signal priority and temporal scope
are specified in Appendix~\ref{sec:supp_metrics_details}.

\begin{table*}[!t]
\centering
\caption{
Performance comparison across all RoboFollow scenes.
All metrics are reported as percentages (\%). For each scene and difficulty
level, the best score per metric across all models is in \textbf{bold}. The
\textbf{Avg} block reports each model's mean IS, ES, and CR over the four
scenes (S1--S4) and all difficulty levels; the best model per Avg metric is bolded.
}
\label{tab:1}
\scriptsize
\setlength{\tabcolsep}{3.1pt}
\renewcommand{\arraystretch}{1.08}

\begin{adjustbox}{width=\textwidth}
\begin{tabular}{
@{} l c
*{3}{N}   
*{3}{B}   
*{3}{N}   
*{3}{P}   
*{3}{c}   
@{}
}
\toprule
\rowcolor{HeaderGray}
\multicolumn{1}{c}{} &
\multicolumn{1}{c}{} &
\multicolumn{3}{c}{\textbf{L0}} &
\multicolumn{3}{c}{\textbf{L1}} &
\multicolumn{3}{c}{\textbf{L2}} &
\multicolumn{3}{c}{\textbf{L3}} &
\multicolumn{3}{c}{\textbf{Avg}} \\

\rowcolor{HeaderGray}
\multicolumn{1}{c}{\multirow{-2}{*}{\textbf{Model}}} &
\multicolumn{1}{c}{\multirow{-2}{*}{\textbf{Scene}}} &
\multicolumn{1}{c}{\textbf{IS}} &
\multicolumn{1}{c}{\textbf{ES}} &
\multicolumn{1}{c}{\textbf{CR}} &
\multicolumn{1}{c}{\textbf{IS}} &
\multicolumn{1}{c}{\textbf{ES}} &
\multicolumn{1}{c}{\textbf{CR}} &
\multicolumn{1}{c}{\textbf{IS}} &
\multicolumn{1}{c}{\textbf{ES}} &
\multicolumn{1}{c}{\textbf{CR}} &
\multicolumn{1}{c}{\textbf{IS}} &
\multicolumn{1}{c}{\textbf{ES}} &
\multicolumn{1}{c}{\textbf{CR}} &
\multicolumn{1}{c}{\textbf{IS}} &
\multicolumn{1}{c}{\textbf{ES}} &
\multicolumn{1}{c}{\textbf{CR}} \\
\midrule

\multirow{4}{*}{\textbf{PI05}}
& S1 & 99.1 & 98.5 & 98.8 & \textbf{45.5} & \textbf{43.8} & \textbf{42.5} & 58.0 & \textbf{58.0} & 58.1 & 44.9 & 43.8 & 40.1 & \multicolumn{1}{c}{\multirow{4}{*}{\textbf{55.7}}} & \multicolumn{1}{c}{\multirow{4}{*}{\textbf{53.0}}} & \multicolumn{1}{c}{\multirow{4}{*}{\textbf{47.3}}} \\
& S2 & \textbf{100.0} & \textbf{100.0} & \textbf{100.0} & \textbf{77.5} & \textbf{69.4} & \textbf{68.8} & 32.4 & 24.6 & 20.6 & 34.2 & 26.7 & 16.7 &  &  &  \\
& S3 & 85.0 & 85.0 & 81.3 & 66.9 & 66.5 & 53.8 & 30.0 & 29.8 & 0.0 & 42.5 & 42.5 & 12.5 &  &  &  \\
& S4 & \textbf{84.1} & 78.1 & 79.6 & \textbf{34.0} & \textbf{31.3} & \textbf{33.3} & 33.6 & 30.4 & 30.0 & 22.9 & \textbf{20.4} & \textbf{20.8} &  &  &  \\
\midrule
\multirow{4}{*}{\textbf{\shortstack[l]{PI05\\empty language}}}
& S1 & 16.0 & 16.0 & 6.3 & 0.5 & 0.5 & 0.0 & 16.6 & 16.6 & 6.3 & 17.4 & 16.6 & 6.7 & \multicolumn{1}{c}{\multirow{4}{*}{12.3}} & \multicolumn{1}{c}{\multirow{4}{*}{8.9}} & \multicolumn{1}{c}{\multirow{4}{*}{5.2}} \\
& S2 & 16.6 & 9.4 & 9.4 & 18.5 & 10.6 & 9.6 & 23.1 & 14.0 & 11.2 & 19.3 & 13.9 & 12.9 &  &  &  \\
& S3 & 6.4 & 2.1 & 0.0 & 8.2 & 4.1 & 0.0 & 6.5 & 2.0 & 0.0 & 7.5 & 2.8 & 0.0 &  &  &  \\
& S4 & 20.4 & 16.5 & 8.2 & 4.7 & 4.3 & 3.3 & 11.0 & 10.4 & 6.4 & 3.7 & 2.4 & 2.5 &  &  &  \\
\midrule
\multirow{4}{*}{\textbf{PI0}}
& S1 & 98.2 & 98.2 & 97.5 & 0.0 & 0.0 & 0.0 & 4.1 & 0.0 & 0.0 & 15.7 & 0.8 & 0.9 & \multicolumn{1}{c}{\multirow{4}{*}{48.4}} & \multicolumn{1}{c}{\multirow{4}{*}{44.9}} & \multicolumn{1}{c}{\multirow{4}{*}{38.9}} \\
& S2 & 95.6 & 95.3 & 95.6 & \textbf{77.5} & 67.1 & 66.3 & 36.3 & 27.4 & 21.9 & \textbf{35.2} & \textbf{29.1} & \textbf{18.8} &  &  &  \\
& S3 & 96.3 & 96.3 & 95.0 & 69.8 & 69.8 & 58.0 & 29.7 & 29.7 & 0.2 & \textbf{45.0} & \textbf{45.0} & \textbf{15.3} &  &  &  \\
& S4 & \textbf{84.1} & \textbf{81.9} & \textbf{81.9} & 26.5 & 23.9 & 20.0 & \textbf{36.6} & \textbf{33.1} & \textbf{32.8} & 23.2 & \textbf{20.4} & 17.5 &  &  &  \\
\midrule
\multirow{4}{*}{\textbf{GROOT N1.6}}
& S1 & 71.4 & 67.1 & 62.5 & 0.0 & 0.0 & 0.0 & 24.0 & 24.0 & 19.4 & 36.0 & 34.0 & 25.8 & \multicolumn{1}{c}{\multirow{4}{*}{32.7}} & \multicolumn{1}{c}{\multirow{4}{*}{25.6}} & \multicolumn{1}{c}{\multirow{4}{*}{20.5}} \\
& S2 & 70.9 & 51.2 & 49.4 & 23.7 & 5.8 & 5.4 & 16.1 & 10.9 & 10.3 & 12.3 & 7.2 & 7.5 &  &  &  \\
& S3 & 56.4 & 52.3 & 48.8 & 33.9 & 30.5 & 19.8 & 23.7 & 17.0 & 0.0 & 20.0 & 14.4 & 0.3 &  &  &  \\
& S4 & 57.8 & 38.2 & 34.1 & 31.9 & 28.7 & 23.3 & 25.3 & 15.8 & 12.4 & 19.6 & 13.0 & 9.2 &  &  &  \\
\midrule
\multirow{4}{*}{\textbf{OpenVLA-OFT}}
& S1 & 9.8 & 9.5 & 6.3 & 0.0 & 0.0 & 0.0 & 2.2 & 2.2 & 0.0 & 2.6 & 1.8 & 1.1 & \multicolumn{1}{c}{\multirow{4}{*}{10.7}} & \multicolumn{1}{c}{\multirow{4}{*}{4.2}} & \multicolumn{1}{c}{\multirow{4}{*}{3.5}} \\
& S2 & 28.9 & 10.3 & 10.6 & 23.3 & 2.1 & 2.5 & 10.2 & 4.3 & 3.8 & 18.9 & 10.3 & 9.2 &  &  &  \\
& S3 & 7.9 & 1.6 & 0.6 & 7.5 & 1.1 & 0.0 & 6.9 & 0.3 & 0.0 & 6.4 & 0.3 & 0.0 &  &  &  \\
& S4 & 23.5 & 9.6 & 7.8 & 9.3 & 6.7 & 6.7 & 5.0 & 0.0 & 0.0 & 8.6 & 7.7 & 7.7 &  &  &  \\
\midrule
\multirow{4}{*}{\textbf{XVLA}}
& S1 & 76.7 & 74.5 & 71.9 & 0.0 & 0.0 & 0.0 & 24.2 & 23.5 & 12.5 & 45.7 & 44.3 & 33.0 & \multicolumn{1}{c}{\multirow{4}{*}{34.3}} & \multicolumn{1}{c}{\multirow{4}{*}{30.9}} & \multicolumn{1}{c}{\multirow{4}{*}{22.3}} \\
& S2 & 76.7 & 65.8 & 66.9 & 6.8 & 6.8 & 0.0 & 38.7 & 30.1 & 24.3 & 16.3 & 10.9 & 7.5 &  &  &  \\
& S3 & 74.0 & 69.9 & 45.6 & 22.5 & 22.2 & 5.0 & 33.0 & \textbf{32.9} & 0.9 & 34.8 & 33.0 & 4.4 &  &  &  \\
& S4 & 38.5 & 34.9 & 39.3 & 6.9 & 5.9 & 3.3 & 29.4 & 23.2 & 25.2 & \textbf{24.5} & 17.2 & 16.7 &  &  &  \\
\midrule
\multirow{4}{*}{\textbf{ACoT-VLA}}
& S1 & \textbf{100.0} & \textbf{100.0} & \textbf{100.0} & 8.7 & 6.2 & 6.3 & \textbf{60.9} & 57.2 & \textbf{59.4} & \textbf{48.0} & \textbf{45.5} & \textbf{42.1} & \multicolumn{1}{c}{\multirow{4}{*}{53.5}} & \multicolumn{1}{c}{\multirow{4}{*}{49.3}} & \multicolumn{1}{c}{\multirow{4}{*}{42.5}} \\
& S2 & 96.4 & 91.6 & 89.4 & 76.3 & 53.4 & 52.5 & \textbf{40.0} & \textbf{33.5} & \textbf{26.6} & 34.0 & 25.0 & 18.3 &  &  &  \\
& S3 & 95.6 & 95.4 & 89.4 & \textbf{77.5} & \textbf{77.0} & \textbf{64.8} & 30.3 & 30.3 & 0.3 & 39.7 & 39.4 & 1.9 &  &  &  \\
& S4 & 78.7 & 72.8 & 72.6 & 23.6 & 23.6 & 23.3 & 26.2 & 22.0 & 18.4 & 20.3 & 15.2 & 15.4 &  &  &  \\
\midrule
\multirow{4}{*}{\textbf{Lingbot-VLA}}
& S1 & 80.9 & 78.4 & 82.5 & 0.0 & 0.0 & 0.0 & 24.2 & 23.7 & 23.8 & 46.4 & 43.5 & 37.2 & \multicolumn{1}{c}{\multirow{4}{*}{35.2}} & \multicolumn{1}{c}{\multirow{4}{*}{32.5}} & \multicolumn{1}{c}{\multirow{4}{*}{27.4}} \\
& S2 & 77.1 & 67.6 & 74.4 & 7.0 & 5.8 & 0.0 & 32.5 & 28.0 & 23.5 & 6.6 & 4.1 & 1.7 &  &  &  \\
& S3 & \textbf{99.1} & \textbf{98.6} & \textbf{97.5} & 24.8 & 24.7 & 9.8 & \textbf{33.2} & 32.6 & 0.2 & 29.8 & 29.2 & 0.1 &  &  &  \\
& S4 & 58.0 & 47.1 & 51.1 & 8.9 & 6.3 & 7.3 & 18.4 & 16.8 & 18.4 & 16.3 & 14.2 & 10.8 &  &  &  \\
\midrule[1pt]
\multirow{4}{*}{\textbf{MOTUS}}
& S1 & 73.4 & 68.8 & 70.6 & 11.3 & 11.1 & 13.8 & 26.9 & 26.1 & 33.1 & 29.8 & 26.7 & 27.0 & \multicolumn{1}{c}{\multirow{4}{*}{36.8}} & \multicolumn{1}{c}{\multirow{4}{*}{30.8}} & \multicolumn{1}{c}{\multirow{4}{*}{27.3}} \\
& S2 & 78.2 & 69.9 & 81.3 & 53.1 & 43.6 & 50.8 & 17.9 & 13.6 & 8.2 & 16.2 & 11.0 & 8.3 &  &  &  \\
& S3 & 67.6 & 61.2 & 45.6 & 42.6 & 37.0 & 17.8 & 30.9 & 26.0 & \textbf{2.0} & 36.6 & 31.8 & 10.3 &  &  &  \\
& S4 & 51.7 & 34.9 & 36.3 & 11.6 & 4.0 & 1.3 & 20.6 & 13.1 & 15.2 & 20.6 & 14.4 & 15.8 &  &  &  \\
\midrule
\multirow{4}{*}{\textbf{FASTWAM}}
& S1 & 83.1 & 83.1 & 94.4 & 0.0 & 0.0 & 0.0 & 13.0 & 4.0 & 3.1 & 41.9 & 36.2 & 31.1 & \multicolumn{1}{c}{\multirow{4}{*}{39.3}} & \multicolumn{1}{c}{\multirow{4}{*}{35.2}} & \multicolumn{1}{c}{\multirow{4}{*}{30.3}} \\
& S2 & 82.5 & 76.5 & 85.0 & 32.0 & 14.1 & 11.7 & 24.8 & 20.4 & 14.6 & 18.0 & 5.3 & 2.9 &  &  &  \\
& S3 & 98.3 & 98.1 & 95.6 & 60.4 & 59.7 & 42.0 & 30.6 & 30.5 & 0.3 & \textbf{45.0} & 43.3 & 4.9 &  &  &  \\
& S4 & 52.1 & 48.4 & 56.3 & 9.1 & 6.9 & 4.0 & 21.1 & 20.6 & 23.2 & 16.7 & 15.5 & 15.4 &  &  &  \\

\bottomrule
\end{tabular}
\end{adjustbox}

\vspace{0.25em}
\footnotesize{
IS: Intent Score; ES: Execution Score; CR: Completion Rate.
Avg: per-model mean of each metric over the four scenes and all difficulty levels.
}
\vspace{-0.5em}
\end{table*}
\section{Experiments and Results}
\label{sec:ex}

In this section, we deploy the RoboFollow benchmark to systematically diagnose the instruction following capabilities of state of the art embodied agents. Rather than merely reporting success rates, we structure our evaluation around several core research questions designed to unmask the illusion of competence.

\subsection{Experimental Setup}

Rather than exhaustively evaluating lower-capacity baselines, we select nine state-of-the-art foundation models spanning the dominant VLA and WAM paradigms. These large-scale models feature extensive pre-training, strong semantic reasoning, and broad community adoption, allowing us to study instruction following across representative pre-trained policies without isolating architecture from data or optimization effects. Our VLA evaluation includes $\pi_0$~\cite{black2024pi_0}, its successor $\pi_{0.5}$~\cite{intelligence2025pi05visionlanguageactionmodelopenworld}, NVIDIA’s GR00T N1.6~\cite{nvidia2025gr00tn1openfoundation}, openvla-oft~\cite{openvla-oft}, xvla~\cite{xvla}, ACoT-VLA~\cite{acot-vla}, and Lingbot-VLA~\cite{lingbot-vla}. For WAM, we evaluate Motus~\cite{bi2025motus} and FAST-WAM~\cite{fastwam}. The four scenes comprise a total of 3,750 training episodes. Our main experiments were all fine-tuned using all the data, while the analysis part of the experiments only used 16 tasks (800 episodes) from scene2 for fine-tuning. Detailed training configurations are provided in the appendix.

\subsection{Main Results}

\textbf{RQ1: Do Current Embodied Agents Genuinely Follow Instructions?} \textit{The short answer is no.} As shown in Tab.~\ref{tab:1}, while leading models project an illusion of competence under in-distribution conditions, their instruction following capabilities collapse precipitously once even minimal perturbations are introduced. Averaged across all scenes, all models show a pronounced decline in IS from L0 to L1–L3, the consistent drop in Intent Score indicates that the degradation cannot be explained solely by low-level execution failures; failures in semantic intent selection are a major contributing factor. On Scenes~1 and~2, which test spatial relation grounding and intrinsic attribute binding, the $\pi$-series models achieve near-saturated L0 performance: $\pi_{0.5}$ obtains 99.1\% and 100.0\% IS, and $\pi_0$ reaches 98.2\% and 95.6\%. However, performance drops sharply at higher levels. In Scene~1, $\pi_0$ falls from 98.2\% at L0 to \textbf{0.0\%} at L1, while $\pi_{0.5}$ declines from 99.1\% to 45.5\% at L1 and 44.9\% at L3. A similar pattern appears in Scene~2, where $\pi_{0.5}$ decreases from 100.0\% at L0 to 34.2\% at L3.

\textbf{RQ2: Where Embodied Agents Break Down?} As shown in Tab.~\ref{tab:1}, an analysis of cross-scene generalization reveals differences across evaluated policies in their generalization over distinct semantic dimensions. Specifically, $\pi_0$, GR00T N1.6, and Motus fail precipitously on the L1 visual grounding evaluation of Scene 1, a pattern consistent with limited spatial grounding and reliance on learned scene--task associations. While $\pi_{0.5}$ exhibits comparatively stronger semantic grounding for these spatial relationships, its performance still degrades substantially under visual perturbations. Interestingly, models such as $\pi_0$ and Motus demonstrate significantly more robust semantic grounding when processing intrinsic object attributes in Scene 2 than they do with the extrinsic spatial relations in Scene 1. 
Failure cases are shown in Appendix~\ref{failure cases}.

\paragraph{Real-robot pilot.}
We evaluate $\pi_{0.5}$ using eight training instructions and eight held-out instructions over the same object set. Each instruction is tested five times: success decreases from 20/40 (50\%) on training instructions to 6/40 (15\%) on held-out instructions. This preliminary comparison uses different instruction sets and pick/stack compositions, rather than matched task pairs; full instructions and counts appear in Appendix~\ref{sec:real_details}.

\section{Analysis}

The severe performance degradation across generalization levels L1–L3 strongly suggests that existing models treat language instructions as shallow task identifiers rather than genuinely understanding their semantics and grounding them to target objects and actions. To investigate the mechanisms of these failures, we conduct an in-depth diagnostic analysis on Scene~2, examining deficiencies in the underlying VLM and evaluating whether recent mitigation strategies, such as stronger VLM backbones and QA co-training, can address this bottleneck.

\subsection{VLM Deficits and The Comprehension-Execution Gap}
\label{subsec:vlm_bottleneck}
A natural first question is whether instruction-following failures originate from the vision-language backbone itself or emerge downstream during action generation. We design a diagnostic protocol that independently probes (i) the VLM's scene comprehension and (ii) the fidelity with which comprehended semantics propagate to action generation.

We constructed a visual QA probe targeting object identities, colors, and spatial relations in Scene~2. The complete set of 20 questions is provided in Appendix~\ref{qa-20 details}. We evaluated the base PaliGemma, the pre-trained $\pi_{0.5}$ and the fine-tuned $\pi_{0.5}$ backbone. Scene understanding accuracy remained consistently low with scores of 1/20, 2/20, and 3/20, respectively.


Even for questions that the fine-tuned VLM answers correctly, about half of these episodes still result in incorrect manipulation behavior. This reveals a two-layered failure structure: the VLM backbone frequently lacks sufficient scene understanding, and even when comprehension succeeds, the action generation pipeline fails to faithfully translate it into behavior. Notably, in Motus, whose VLM backbone is entirely frozen throughout training, we observe a comparable disconnect between correct visual comprehension and successful manipulation execution. A natural follow-up question is whether these bottlenecks can be alleviated by (i) employing a substantially more powerful VLM backbone, and (ii) applying recently proposed optimization strategies specifically designed to preserve or enhance instruction-following capabilities. 


\subsection{Can Stronger VLMs and Existing Optimizations Help?}
\label{subsec:methods}

\begin{table*}[!htbp]
\centering
\caption{Performance comparison on Scene 2 (all metrics in \%).}
\label{tab:analysis_scene2}
\scriptsize
\setlength{\tabcolsep}{3.2pt}
\renewcommand{\arraystretch}{1.10}

\begin{threeparttable}
\begin{adjustbox}{width=\textwidth}
\begin{tabular}{
@{} >{\raggedright\arraybackslash}p{2.9cm}
*{3}{N}
*{3}{B}
*{3}{N}
*{3}{P}
@{}
}
\toprule
\rowcolor{HeaderGray}
\multicolumn{1}{c}{} &
\multicolumn{3}{c}{\textbf{L0}} &
\multicolumn{3}{c}{\textbf{L1}} &
\multicolumn{3}{c}{\textbf{L2}} &
\multicolumn{3}{c}{\textbf{L3}} \\

\rowcolor{HeaderGray}
\multicolumn{1}{c}{\multirow{-2}{*}{\textbf{Model}}} &
\multicolumn{1}{c}{\textbf{IS}} &
\multicolumn{1}{c}{\textbf{ES}} &
\multicolumn{1}{c}{\textbf{CR}} &
\multicolumn{1}{c}{\textbf{IS}} &
\multicolumn{1}{c}{\textbf{ES}} &
\multicolumn{1}{c}{\textbf{CR}} &
\multicolumn{1}{c}{\textbf{IS}} &
\multicolumn{1}{c}{\textbf{ES}} &
\multicolumn{1}{c}{\textbf{CR}} &
\multicolumn{1}{c}{\textbf{IS}} &
\multicolumn{1}{c}{\textbf{ES}} &
\multicolumn{1}{c}{\textbf{CR}} \\
\midrule

pi05
& 93.6 & 89.0 & 89.4
& 75.7 & 42.0 & 40.4
& 37.8 & 30.3 & 25.7
& 45.8 & 24.9 & 19.2 \\

pi05\_cfg\_1.2
& 55.9 & 44.7 & 46.9
& 50.4 & 25.0 & 25.0
& 31.2 & 22.4 & 21.3
& 31.2 & 14.2 & 8.3 \\

pi05\_cfg\_1.5
& 43.6 & 25.9 & 25.0
& 37.8 & 23.5 & 25.0
& 28.6 & 20.0 & 20.6
& 24.6 & 12.3 & 10.4 \\

qwengroot
& 67.9 & 63.1 & 65.6
& 6.3 & 0.0 & 0.0
& 31.1 & 21.7 & 17.2
& 14.8 & 1.3 & 1.7 \\

qwengroot\_qa1
& 60.1 & 46.4 & 45.0
& 3.5 & 0.8 & 0.8
& 26.4 & 15.6 & 10.3
& 18.9 & 8.8 & 9.2 \\

qwengroot\_qa2
& 68.2 & 61.3 & 60.6
& 0.0 & 0.0 & 0.0
& 21.8 & 16.0 & 16.6
& 0.0 & 0.0 & 0.0 \\

qwengroot\_qa3
& 66.6 & 50.5 & 46.3
& 2.9 & 0.0 & 0.0
& 30.8 & 19.3 & 15.0
& 0.0 & 0.0 & 0.0 \\

langforce
& 65.1 & 44.2 & 35.0
& 6.3 & 4.0 & 0.0
& 18.6 & 14.7 & 14.1
& 0.0 & 0.0 & 0.0 \\

\bottomrule
\end{tabular}
\end{adjustbox}

\vspace{0.35em}
\begin{tablenotes}[flushleft]
\footnotesize
\item IS: Intent Score; ES: Execution Score; CR: Completion Rate.
\item All models are fine-tuned only on Scene 2 data.
\item In pi05-cfg-x, x denotes the CFG guidance scale.
\item qa1 uses ShareGPT4V-COCO QA, qa2 uses our Scene 2 QA dataset, and qa3 uses mixed QA.
\end{tablenotes}
\end{threeparttable}

\vspace{-0.6em}
\end{table*}

The analysis in Sec.~\ref{subsec:vlm_bottleneck} identifies two cascading bottlenecks: inadequate scene comprehension in the VLM backbone, and a comprehension-to-execution gap in the action head. We now ask whether three representative mitigation strategies can alleviate these failures: (i)~upgrading the VLM backbone, (ii)~QA co-training to preserve linguistic competence, and (iii)~language-conditioned guidance. Results are summarized in Table~\ref{tab:analysis_scene2}.

Following the above observations, we evaluated Qwen3-VL (4B), a significantly more powerful vision-language backbone. On our fine-grained QA probe test set for Scene 2, Qwen3-VL-4B achieved an impressive success rate of 19/20, demonstrating highly robust zero-shot spatial and attribute comprehension. 

\ding{170} We trained Qwen-GR00T by integrating the Qwen3-VL-4B backbone with the GR00T diffusion action head, to mitigate the widely observed phenomenon of catastrophic forgetting, where action fine-tuning erases pre-trained linguistic capabilities, we implemented a QA co-training strategy. During this process, visual QA pairs and manipulation trajectories were jointly optimized. However, the out-of-distribution instruction following capabilities across L1–L3 still exhibit a catastrophic collapse in high-entropy scenarios. This indicates that merely possessing a stronger VLM and explicitly maintaining its QA capabilities during training is insufficient to bridge the comprehension to execution gap when complex physical grounding is required.

We further examined two strategies that aim to amplify linguistic influence over generated actions.

\ding{170} \textit{LangForce}~\cite{lian2026langforce}, a recent method that strengthens text–action correlation during training, yields only marginal changes. We attribute this to the method's underlying mechanism: while LangForce successfully amplifies the statistical correlation between the instruction text and the generated motion distribution, it lacks explicit supervision for genuine, compositional semantic understanding.

\ding{170} \textit{Classifier-Free Guidance} (CFG)~\cite{zhan2026stable}, a standard technique for boosting conditioning signals in diffusion policies, proves counterproductive in our setting. Increasing the guidance scale to 1.2 and 1.5 degrades even L0 performance and produces erratic trajectories. In low-entropy benchmarks, the unconditional prediction provides a stable baseline because vision alone largely determines the task; in our high-entropy scenes, dropping the language input forces the model into multimodal guessing, rendering the CFG residual dominated by noise rather than a clean semantic signal. 

None of the evaluated strategies, whether targeting the VLM backbone, the training objective, or the inference procedure, yields satisfactory generalization beyond L0 in high-entropy environments. These results leave instruction following unresolved in the tested configurations, without ruling out improvements from broader data distributions or alternative training procedures. These observations motivate increasing structural scene and semantic diversity while preserving held-out combinations, and supervising grounded language-to-action alignment during adaptation. These directions remain hypotheses for future evaluation.

\subsection{Additional Fine-Tuning Controls}
\label{sec:control_details}
We examine whether the observed generalization gaps persist under
changes to the fine-tuning setup. These controls use $\pi_{0.5}$ on the 16 Scene~2 tasks, separately from the full-benchmark experiment. Table~\ref{tab:training_controls} reports the Intent Scores under changes in demonstration count, instruction variants, and training duration.
\begin{table}[H]
\centering
\caption{Scene~2 fine-tuning controls: Intent Score (\%). Each panel varies the indicated factor. 1/3/5 denotes total instruction variants.}
\label{tab:training_controls}
\small
\begin{tabular}{llrrrr}
\toprule
Factor & Value & L0 & L1 & L2 & L3 \\
\midrule
Demonstrations/task & 25 & 88.4 & 69.8 & 33.5 & 39.8 \\
& 50 & 93.6 & 75.7 & 37.8 & 45.8 \\
\midrule
Instruction variants & 1 & 93.1 & 72.8 & 31.6 & 41.5 \\
& 3 & 93.6 & 75.7 & 37.8 & 45.8 \\
& 5 & 93.1 & 79.0 & 32.0 & 53.0 \\
\midrule
Training steps & 2k & 86.4 & 79.2 & 31.5 & 38.9 \\
& 4k & 92.4 & 77.6 & 35.7 & 47.6 \\
& 6k & 93.6 & 75.7 & 37.8 & 45.8 \\
\bottomrule
\end{tabular}
\end{table}

Doubling demonstrations from 25 to 50 per task improves all Intent Scores, but leaves large gaps: L0--L2 changes from 54.9 to 55.8 percentage points, and L0--L3 from 48.6 to 47.8.

Increasing instruction variants from one to five improves L1 (72.8\% to 79.0\%) and L3 (41.5\% to 53.0\%), while L2 varies non-monotonically (31.6\%, 37.8\%, 32.0\%). From 2k to 6k steps, L0/L2 improve and L1 decreases; L3 peaks at 4k. At 6k, L0--L2/L3 gaps remain 55.8/47.8 points.

Within the tested ranges, these controls improve $\pi_{0.5}$ on Scene~2 but leave substantial L0--L2/L3 gaps. Structural layout diversity remains untested; evaluating it requires new training layouts while keeping evaluation layouts held out. Training on existing L1 test layouts would invalidate the split.
\section{Conclusion}

In this work, we introduced RoboFollow, a diagnostic benchmark for evaluating whether embodied agents genuinely follow linguistic instructions. RoboFollow combines high-entropy scene design, a hierarchical L0--L3 protocol, and decoupled intent-execution scoring to diagnose instruction following across spatial, attribute-based, procedural, and logical semantics. Under our fine-tuning setup, experiments on the evaluated VLA and WAM policies show that performance degrades sharply under
minimal visual and semantic perturbations, and that existing optimization strategies remain insufficient. These results identify robust instruction following as a critical bottleneck for controllable and reliable embodied agents.

\section{Limitations}

RoboFollow is intended as a diagnostic benchmark rather than a comprehensive test of all embodied capabilities. Its controlled high-entropy scenes simplify object geometry, task horizons, and interaction dynamics to isolate instruction-following failures from execution confounds. Therefore, it does not fully cover long-horizon planning, contact-rich manipulation, open-vocabulary diversity, or large-scale real-world deployment. Our controls vary demonstration count, instruction variants, and training duration, but do not test increased structural layout diversity during training. The real-robot pilot compares different instruction sets, and we have not established transfer across simulators or to more complex real-world tasks. Future work should extend RoboFollow to richer embodiments and real-world scenarios while preserving its language-necessary design principle.

\clearpage
\acknowledgments{We thank the anonymous reviewers and the area chair for their
constructive feedback and helpful suggestions. This work was supported in part by the Natural Science Foundation of China (Grant No.62503323).}



\bibliography{example}  

\clearpage
\appendix
\section{Benchmark Details}
\label{sec:supp_benchmark_details}

RoboFollow is built upon the RoboTwin2.0 simulation platform\cite{chen2025robotwin20scalabledata}. We define four scenes and carefully partition the training data and evaluation tasks. In the training data, each task is provided with 50 episodes. When the relative object arrangements remain identical across episodes, a slight random positional perturbation of 1–2 cm is applied to each object. Furthermore, the instruction for each training episode is randomly sampled from a set of paraphrased variants to enrich semantic diversity during training. The template for the training instructions for scene1 is shown in Tab.\ref{tab:scene1_train_instructions}. The four scenes comprise a total of 3,750 training episodes. Our main experiments were all fine-tuned using all the data, while the analysis part of the experiments only used 16 tasks (800 episodes) from scene2 for fine-tuning.

\begin{small}
\setlength{\LTcapwidth}{\textwidth}
\begin{longtable}{@{}c p{0.88\textwidth}@{}}
\caption{Scene 1 training instructions.}
\label{tab:scene1_train_instructions} \\
\toprule
\rowcolor{headerblue}
\textbf{ID} & \textbf{Instruction Templates} \\
\midrule
\endfirsthead
\toprule
\rowcolor{headerblue}
\textbf{ID} & \textbf{Instruction Templates} \\
\midrule
\endhead
\midrule
\multicolumn{2}{r}{\textit{Continued on next page}} \\
\endfoot
\bottomrule
\endlastfoot

\scenerow{Default scene}
1
& \textbf{T0:} Use your left arm to pick up the block to the right of the red ball. \newline
  \textbf{T1:} Use your left arm to grasp the block to the right of the red ball. \newline
  \textbf{T2:} Pick up the block to the right of the red ball using your left arm. \\
\rowcolor{lightgray}
2
& \textbf{T0:} Use your right arm to pick up the block to the right of the red ball. \newline
  \textbf{T1:} Use your right arm to grasp the block to the right of the red ball. \newline
  \textbf{T2:} Pick up the block to the right of the red ball using your right arm. \\

3
& \textbf{T0:} Use your left arm to pick up the block to the left of the green ball. \newline
  \textbf{T1:} Use your left arm to grasp the block to the left of the green ball. \newline
  \textbf{T2:} Pick up the block to the left of the green ball using your left arm. \\
\rowcolor{lightgray}
4
& \textbf{T0:} Use your right arm to pick up the block in front of the red cylinder. \newline
  \textbf{T1:} Use your right arm to grasp the block in front of the red cylinder. \newline
  \textbf{T2:} Pick up the block in front of the red cylinder using your right arm. \\

5
& \textbf{T0:} Use your right arm to pick up the block to the right of the red ball and place it behind the red cylinder. \newline
  \textbf{T1:} Use your right arm to grasp the block to the right of the red ball and put it behind the red cylinder. \newline
  \textbf{T2:} Pick up the block to the right of the red ball and place it behind the red cylinder using your right arm. \\
\rowcolor{lightgray}
6
& \textbf{T0:} Use your right arm to pick up the block to the right of the red ball and place it to the right of the red cylinder. \newline
  \textbf{T1:} Use your right arm to grasp the block to the right of the red ball and put it to the right of the red cylinder. \newline
  \textbf{T2:} Pick up the block to the right of the red ball and place it on the right side of the red cylinder using your right arm. \\

7
& \textbf{T0:} Use your left arm to pick up the block to the right of the red ball and place it behind the red ball. \newline
  \textbf{T1:} Use your left arm to grasp the block to the right of the red ball and put it behind the red ball. \newline
  \textbf{T2:} Pick up the block to the right of the red ball and place it behind the red ball using your left arm. \\
\rowcolor{lightgray}
8
& \textbf{T0:} Use your left arm to pick up the block to the right of the red ball and place it to the left of the red ball. \newline
  \textbf{T1:} Use your left arm to grasp the block to the right of the red ball and put it to the left of the red ball. \newline
  \textbf{T2:} Pick up the block to the right of the red ball and place it on the left side of the red ball using your left arm. \\

9
& \textbf{T0:} Use your left arm to pick up the block to the left of the green ball and place it behind the green ball. \newline
  \textbf{T1:} Use your left arm to grasp the block to the left of the green ball and put it behind the green ball. \newline
  \textbf{T2:} Pick up the block to the left of the green ball and place it behind the green ball using your left arm. \\
\rowcolor{lightgray}
10
& \textbf{T0:} Use your left arm to pick up the block to the left of the green ball and place it in front of the green ball. \newline
  \textbf{T1:} Use your left arm to grasp the block to the left of the green ball and put it in front of the green ball. \newline
  \textbf{T2:} Pick up the block to the left of the green ball and place it in front of the green ball using your left arm. \\

11
& \textbf{T0:} Use your left arm to pick up the block to the left of the green ball and place it behind the red ball. \newline
  \textbf{T1:} Use your left arm to grasp the block to the left of the green ball and put it behind the red ball. \newline
  \textbf{T2:} Pick up the block to the left of the green ball and place it behind the red ball using your left arm. \\
\rowcolor{lightgray}
12
& \textbf{T0:} Use your left arm to pick up the block to the left of the green ball and place it to the left of the red ball. \newline
  \textbf{T1:} Use your left arm to grasp the block to the left of the green ball and put it to the left of the red ball. \newline
  \textbf{T2:} Pick up the block to the left of the green ball and place it on the left side of the red ball using your left arm. \\

13
& \textbf{T0:} Use your right arm to pick up the block in front of the red cylinder and place it behind the red cylinder. \newline
  \textbf{T1:} Use your right arm to grasp the block in front of the red cylinder and put it behind the red cylinder. \newline
  \textbf{T2:} Pick up the block in front of the red cylinder and place it behind the red cylinder using your right arm. \\
\rowcolor{lightgray}
14
& \textbf{T0:} Use your right arm to pick up the block in front of the red cylinder and place it to the right of the red cylinder. \newline
  \textbf{T1:} Use your right arm to grasp the block in front of the red cylinder and put it to the right of the red cylinder. \newline
  \textbf{T2:} Pick up the block in front of the red cylinder and place it on the right side of the red cylinder using your right arm. \\

15
& \textbf{T0:} Use your right arm to pick up the block in front of the red cylinder and place it behind the green ball. \newline
  \textbf{T1:} Use your right arm to grasp the block in front of the red cylinder and put it behind the green ball. \newline
  \textbf{T2:} Pick up the block in front of the red cylinder and place it behind the green ball using your right arm. \\
\rowcolor{lightgray}
16
& \textbf{T0:} Use your right arm to pick up the block in front of the red cylinder and place it in front of the green ball. \newline
  \textbf{T1:} Use your right arm to grasp the block in front of the red cylinder and put it in front of the green ball. \newline
  \textbf{T2:} Pick up the block in front of the red cylinder and place it in front of the green ball using your right arm. \\

\end{longtable}
\end{small}

The following are the specific training and testing task designs for the four scenarios:


\textbf{Scene 1: Extrinsic Spatial Relations.}

Variants of the Scene 1 Layout are shown in Tab.\ref{tab:scene1_variants}:

\begin{small}
\setlength{\LTcapwidth}{\textwidth}
\begin{longtable}{@{}>{\centering\arraybackslash}p{1.8cm} p{\dimexpr\textwidth-1.8cm-2\tabcolsep\relax}@{}}
\caption{Scene 1 Layout Variants from the Default Scene}
\label{tab:scene1_variants} \\
\toprule
\endfirsthead

\toprule
\endhead

\bottomrule
\endlastfoot

Scene 1.1 & Swaps the green ball and the red cylinder, and then swaps the red cylinder and the red ball. \\
Scene 1.2 & Swaps the red ball and the green ball. \\
Scene 1.3 & Swaps the green ball and the red cylinder. \\

\end{longtable}
\end{small}

\begin{figure}[!ht]
  \centering
  \includegraphics[width=\linewidth]{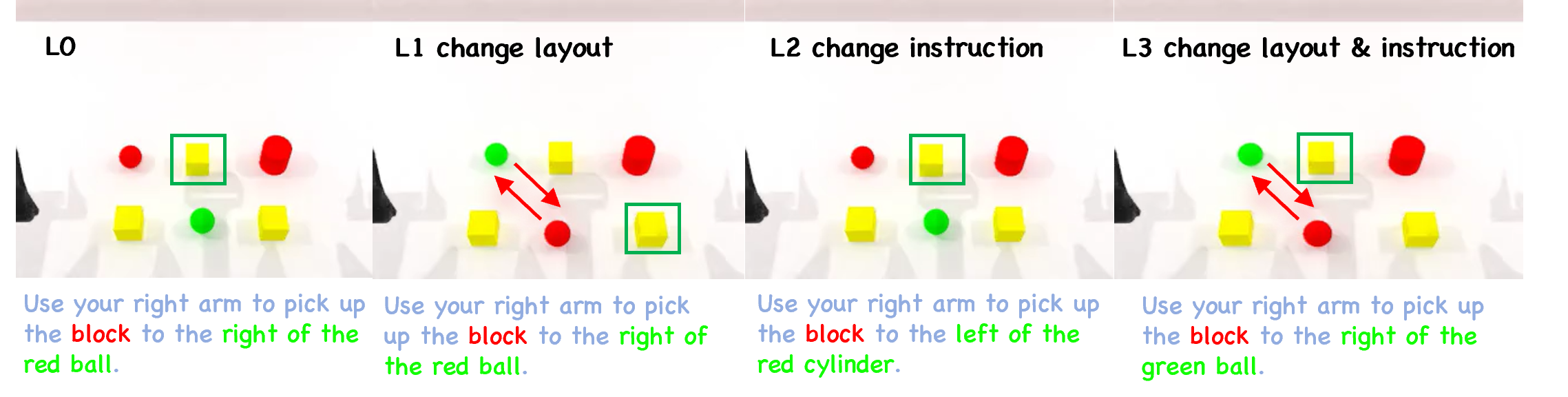}
  \caption{Example of Scene 1.}
  \label{fig:scene1}
\end{figure}

In this example shown in Fig.~\ref{fig:scene1}, the scene–instruction pair in L0 appears in the training set. In L1, the instruction from L0 is retained, but the scene layout is changed, where the positions of the red ball and the green ball are swapped. In this setting, we expect the model to preserve the correct understanding of the phrase ``to the right of the red ball''. Specifically, the model should grasp the block located to the right of the relocated red ball, rather than relying on memorized coordinates from the L0 training examples and incorrectly associating ``to the right of the red ball'' with a fixed spatial position (e.g., the upper-center region of the image).

In L2, the scene layout remains the same as the training-time default scene. The target block is still located at the upper-center position, but the instruction given to the model is changed to ``the block to the left of the red cylinder.'' This setting tests whether the model can correctly interpret a novel compositional instruction. Although the individual concepts such as left and red cylinder appear in the training data, their combination does not.

Finally, L3 simultaneously changes both the scene layout and the instruction. In this case, the layout is switched, and the instruction becomes ``the block to the right of the green ball.'' Similar to L2, the individual concepts (right, green ball) have appeared in the training data, but their composition has not.

Importantly, the target actions in all these cases are present in the training set. Therefore, any failure cannot be attributed to the model encountering unseen actions, but rather reflects its ability to generalize across novel scene configurations and compositional instructions.

\vspace{6pt}

\textbf{Scene 2: Intrinsic Object Properties.}

Variants of the Scene 2 Layout are shown in Tab.\ref{tab:scene2_variants}:

\begin{small}
\setlength{\LTcapwidth}{\textwidth}
\begin{longtable}{@{}>{\centering\arraybackslash}p{1.8cm} p{\dimexpr\textwidth-1.8cm-2\tabcolsep\relax}@{}}
\caption{Scene 2 Layout Variants from the Default Scene}
\label{tab:scene2_variants} \\
\toprule
\endfirsthead

\toprule
\endhead

\bottomrule
\endlastfoot

Scene 2.1 & Swaps the large red cylinder and the large blue cube. \\
Scene 2.2 & Swaps the large blue cube and the small blue cylinder. \\
Scene 2.3 & Swaps the small red cube and the small blue cylinder. \\
Scene 2.4 & Swaps the large red cylinder and the small red cube. \\
Scene 2.5 & Swaps the green bowl and the yellow bowl. \\
Scene 2.6 & Swaps the small red cube and the small blue cylinder, and also swaps the green bowl and the yellow bowl. \\

\end{longtable}
\end{small}

\begin{figure}[!ht]
  \centering
  \includegraphics[width=\linewidth]{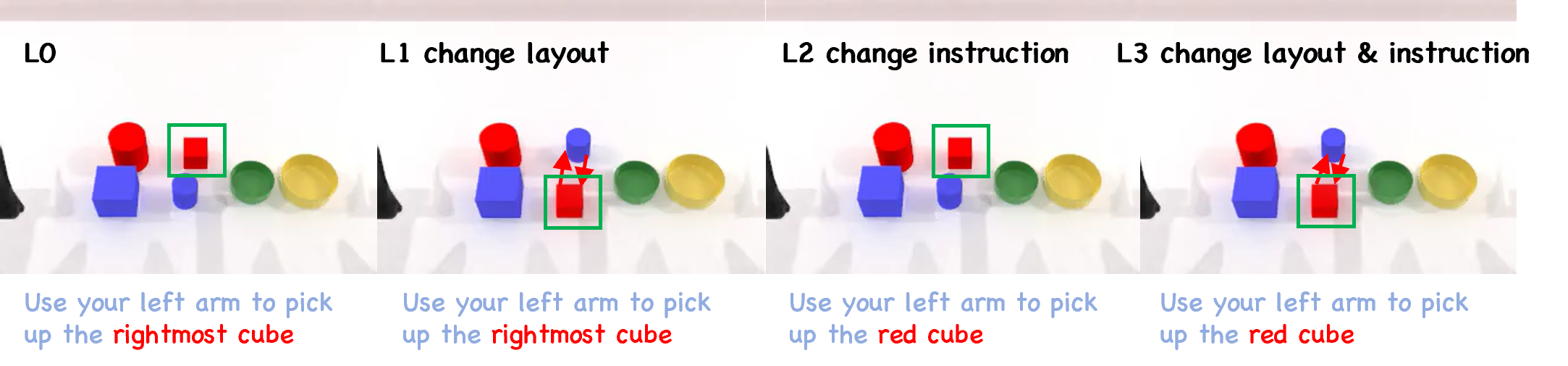}
  \caption{Example of Scene 2.}
  \label{fig:scene2}
\end{figure}

\clearpage
In this example shown in Fig.~\ref{fig:scene2}, L1 retains the instruction from L0, but the scene layout is modified by swapping the positions of the blue cylinder and the red cube. Under this setting, we expect the model to preserve the semantic understanding of ``rightmost cube'' and correctly grasp the red cube at its new rightmost position, rather than memorizing the specific coordinates observed during training and associating ``rightmost cube'' with a fixed location in the training scene.

In L2, the scene layout remains the default scene observed during training, where the target object is still the small red cube. However, the instruction given to the model is modified to refer explicitly to the ``red cube''. This setting evaluates whether the model can correctly interpret the new instruction. Notably, the concepts ``red'' and ``cube'' both appear in the training data, but the specific compositional phrase ``red cube'' does not, allowing us to test the model’s ability to generalize compositionally.

Finally, L3 introduces changes to both the scene layout and the instruction. Importantly, the target actions in all these settings appear in the training dataset, ensuring that any execution failures cannot be attributed to the model encountering previously unseen actions, but rather reflect its capability in instruction understanding and generalization.

\vspace{6pt}

\textbf{Scene 3: Fine grained Action Modulation and Trajectory Constraints.}

Variants of the Scene 3 Layout are shown in Tab.\ref{tab:scene3_variants}:
\vspace{6pt}

\begin{small}
\setlength{\LTcapwidth}{\textwidth}
\begin{longtable}{@{}>{\centering\arraybackslash}p{1.8cm} p{\dimexpr\textwidth-1.8cm-2\tabcolsep\relax}@{}}
\caption{Scene 3 Layout Variants from the Default Scene}
\label{tab:scene3_variants} \\
\toprule
\endfirsthead

\toprule
\endhead

\bottomrule
\endlastfoot

Scene 3.1 & Swaps the green slab and the red slab. \\
Scene 3.2 & Swaps the blue slab and the yellow slab. \\
Scene 3.3 & Swaps the red slab and the yellow slab. \\
Scene 3.4 & Swaps the green slab and the blue slab. \\

\end{longtable}
\end{small}

\begin{figure}[!ht]
  \centering
  \includegraphics[width=\linewidth]{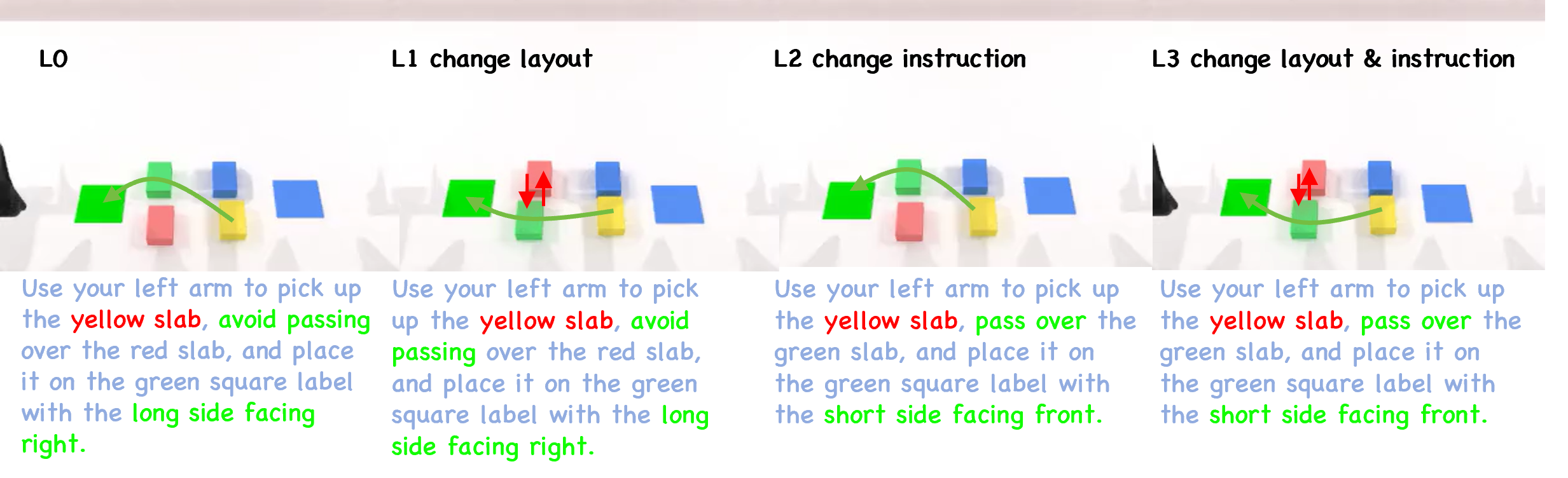}
  \caption{Example of Scene 3.}
  \label{fig:scene3}
\end{figure}

In this example shown in Fig.~\ref{fig:scene3}, although L1 retains the same instruction as L0, the scene layout is modified by swapping the positions of the green slab and the red slab. Under this change, we expect the model to preserve its correct interpretation of the constraint ``avoid passing over the red slab''. Specifically, the model should move the yellow slab along a trajectory that passes over the green slab, while still avoiding the red slab in the new location.

In L2, the scene retains the default training layout, while the instruction is reformulated. In this example, ``avoid passing over the red slab'' is replaced with ``pass over the green slab'', and ``long side facing right'' is replaced with ``short side facing front''. Under the benchmark’s paired-route convention, the route expressions specify the same intended route, and the orientation expressions describe the same final placement orientation. Thus, L2 changes the linguistic formulation while preserving the intended behavior. All constituent concepts are present in the training instructions. 

Finally, L3 simultaneously modifies both the scene layout and the instruction, combining the changes introduced in L1 and L2.

\vspace{6pt}

\textbf{Scene 4: Elementary Logical Grounding.}

Variants of the Scene 4 Layout are shown in Tab.\ref{tab:scene4_variants}:

\begin{small}
\setlength{\LTcapwidth}{\textwidth}
\begin{longtable}{@{}>{\centering\arraybackslash}p{1.8cm} p{\dimexpr\textwidth-1.8cm-2\tabcolsep\relax}@{}}
\caption{Scene 4 Layout Variants from the Default Scene}
\label{tab:scene4_variants} \\
\toprule
\endfirsthead

\toprule
\endhead

\bottomrule
\endlastfoot

Scene 4.1 & Removes the cube inside the bowl and swaps the blue sphere and the red sphere. \\
Scene 4.2 & Removes the cube inside the bowl, places the blue cube on top of the yellow cylinder, moves the green cylinder down to the tabletop at the original blue-cube position, and moves the red sphere to the right side\\
Scene 4.3 & Removes the cube inside the bowl. \\
Scene 4.4 & Swaps the blue sphere and the red sphere. \\
Scene 4.5 & Places the green cylinder on top of the yellow cylinder. \\
Scene 4.6 & Swaps the positions of the blue cube (together with the green cylinder stacked on it) and the yellow cylinder. \\
Scene 4.7 & Moves the red sphere to the right side. \\

\end{longtable}
\end{small}

\begin{figure}[!ht]
  \centering
  \includegraphics[width=\linewidth]{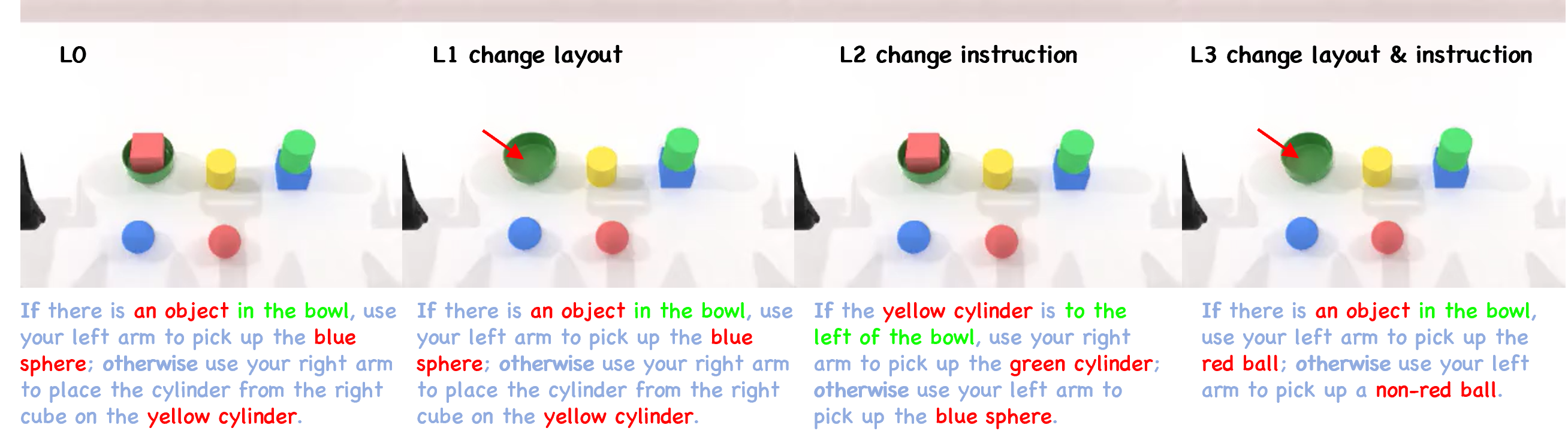}
  \caption{Example of Scene 4.}
  \label{fig:scene4}
\end{figure}

In this example shown in Fig.~\ref{fig:scene4}, L1 retains the same instruction as L0, but the scene layout is modified: the object inside the bowl is removed. Under this change, we expect the model to correctly interpret the semantic condition and execute the otherwise branch of the instruction.

In L2, the scene layout remains the default scene observed during training, while the instruction provided to the model is altered by modifying the condition. This setting evaluates whether the model can correctly understand and follow the updated instruction.

Finally, L3 introduces changes to both the scene layout and the instructional execution branch, simultaneously testing the model’s ability to generalize across variations in both environmental configuration and instruction semantics.

\subsection{Scene Entropy Computation}
\label{sec:entropy_details}
Group training tasks by their task-independent initial scene specification: fixtures, visible objects and attributes, placement distributions, and observable initial-state predicates. Exclude task names, language, and goal annotations; random jitters from a shared distribution and paraphrases do not create new scene groups or task labels. Scene~1/2/3 each has one group of 16 task labels; Scene~4 has three groups of 16, 7, and 4. With 50 demonstrations per task,
\[
H_{\mathrm{scene}}=\frac{64}{75}\log_2 16+
\frac{7}{75}\log_2 7+\frac{4}{75}\log_2 4=3.782\ \mathrm{bits}.
\]
For example, a 16-task group contributes $(16/75)\log_2 16$ bits. The corresponding LIBERO Spatial/Object/Goal/Long values are $0/0/3.322/0.200$ bits, whose equally weighted mean is $0.880$ bits.


\section{Metrics Details}
\label{sec:supp_metrics_details}

We evaluate each episode using a stage-based protocol rather than a single binary success label. Across scenes, a task is decomposed into up to three functionally defined stages. Stage 1 evaluates initial source-object grounding (or initial contact for push actions), Stage 2 evaluates the action-specific manipulation objective, and Stage 3 evaluates the post-manipulation outcome, such as retraction quality or final placement pose. For pickup-only tasks that do not involve an intermediate manipulation objective, Stage 2 is omitted. This design allows the evaluator to capture whether a policy fails at early grounding, during manipulation, or after the main action has nominally been completed.

For each applicable stage, we report two complementary metrics. Intent Score measures whether the policy grounds the instruction to the correct object, relation, or logical branch and moves in the semantically appropriate direction, even when the physical execution is imperfect. Execution Score measures whether the corresponding physical sub-goal is successfully completed. This separation is intended to make semantic misgrounding and execution failure more distinguishable than in conventional final-state success evaluation. As a result, approaching the correct target without completing the manipulation can still receive partial semantic credit, while interacting with an incorrect object is penalized even if the final state happens to look plausible.

The stage definitions follow a shared template across scenes while remaining aligned with the semantic focus of each scenario. In Scenes 1 and 2, the evaluation  focuses on correct source selection, completion of the specified action objective, and post-action stabilization or retraction. In Scene 3, the intermediate stage focuses on compliance with trajectory constraints, and the final stage evaluates placement quality and orientation. In Scene 4, the same framework is extended to instructions involving negation, conditionals, and temporal order, with stage definitions adapted to standard, conditional, and sequential tasks. 

CR uses action-dependent criteria: pickup-like tasks require both final
IS and ES to reach 0.8; other tasks use target-layout or action-progress
signals, with designated execution-stage scores as fallbacks.
In Scenes~1, 2, and~4, \texttt{stage3\_finish} is not checked separately
by CR, but its credit remains in the totals used for pickup-like tasks.
Scene~3 uses positive execution credit in its final placement/orientation
stage. CR does not require full credit at every intermediate stage,
and subsequent actions can affect CR by changing the evaluated final state.


\section{Policies Training Details}
\label{policies training details}

In the main experiments, each policy was fine-tuned on the full set of 3,750 training episodes. Training steps were selected per model based on in-distribution convergence, ensuring each policy reached its performance plateau before evaluation.

Table~\ref{tab:training_details} summarizes the batch size and training duration for each policy.

\begin{table}[H]
\centering
\caption{Training hyperparameters for each policy. All models are fine-tuned on the full set of 3,750 training episodes. Training steps are selected per model based on in-distribution (L0) convergence.}
\label{tab:training_details}
\begin{tabular}{lcc}
\toprule
Model & Batch Size & Steps \\
\midrule
$\pi_{0.5}$    & 32 & 20,000 \\
$\pi_0$        & 32 & 30,000 \\
GR00T N1.6     & 64 & 32,000 \\
Motus          & 32 & 8,000  \\
OpenVLA-OFT    & 32 & 30,000 \\
XVLA           & 32 & 30,000 \\
ACoT-VLA       & 16 & 20,000 \\
Lingbot-VLA    & 32 & 30,000 \\
FAST-WAM       & 64 & 15,000 \\
\bottomrule
\end{tabular}
\end{table}

\subsection{Real-Robot Pilot Details}
\label{sec:real_details}
\begin{figure}[H]
  \centering
  \includegraphics[width=0.5\linewidth]{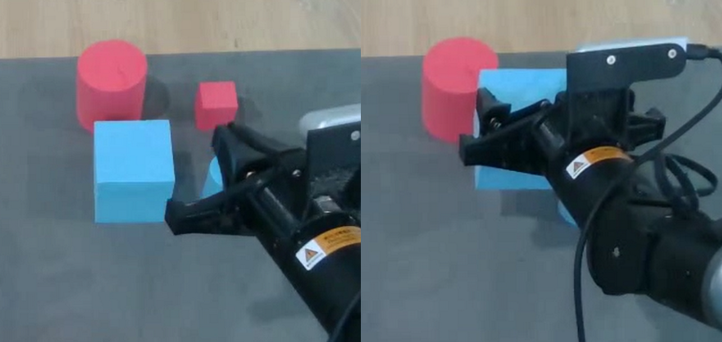}
  \caption{Real-robot experimental setup and representative
  executions of $\pi_{0.5}$ with the right arm.}
  \label{fig:real_robot}
\end{figure}

We use $\pi_{0.5}$, all instructions specify the right arm. The eight training instructions and eight held-out instructions are each evaluated in five trials, giving 40 trials per set and 80 trials in total. These counts refer to evaluation rollouts, not training demonstrations or independent model-training runs. Table~\ref{tab:real_instructions} lists both instruction sets and their per-instruction success counts. Figure~\ref{fig:real_robot} shows the experimental setup and representative real-robot executions.

\begin{small}
\setlength{\LTcapwidth}{\textwidth}
\begin{longtable}{@{}c p{0.72\textwidth} c@{}}
\caption{Real-robot instructions and success counts. Each instruction has five evaluation trials. IDs are local to each set and do not indicate matched pairs.}
\label{tab:real_instructions} \\
\toprule
ID & Instruction & Successes/trials \\
\midrule
\endfirsthead
\toprule
ID & Instruction & Successes/trials \\
\midrule
\endhead
\bottomrule
\endfoot
\multicolumn{3}{l}{\textbf{Training instructions}} \\
1 & Use your right arm to pick up the larger red object. & 5/5 \\
2 & Use your right arm to pick up the leftmost cube. & 3/5 \\
3 & Use your right arm to pick up the rightmost cube. & 5/5 \\
4 & Use your right arm to pick up the blue cylinder. & 1/5 \\
5 & Use your right arm to stack the leftmost cube onto the larger red object. & 2/5 \\
6 & Use your right arm to stack the leftmost cube onto the smaller blue object. & 3/5 \\
7 & Use your right arm to stack the blue cylinder onto the leftmost cube. & 0/5 \\
8 & Use your right arm to stack the blue cylinder onto the rightmost cube. & 1/5 \\
\midrule
\multicolumn{3}{l}{\textbf{Held-out instructions}} \\
1 & Use your right arm to pick up the blue cube. & 4/5 \\
2 & Use your right arm to pick up the larger blue object. & 2/5 \\
3 & Use your right arm to pick up the larger cylinder. & 0/5 \\
4 & Use your right arm to pick up the red cube. & 0/5 \\
5 & Use your right arm to pick up the smaller cube. & 0/5 \\
6 & Use your right arm to pick up the smaller red object. & 0/5 \\
7 & Use your right arm to stack the blue cylinder onto the blue cube. & 0/5 \\
8 & Use your right arm to place the blue cylinder on top of the red cube. & 0/5 \\
\end{longtable}
\end{small}

Success is 20/40 (50\%) on training instructions and 6/40 (15\%) on held-out instructions, a descriptive difference of 35 percentage points. Because every instruction has five trials, the pooled rate equals the unweighted mean of instruction-level rates within each set.

The training set contains four pick and four stack instructions, while the held-out set contains six pick and two stack instructions. Within these categories, success is 14/20 (70\%) versus 6/30 (20\%) for pick instructions, and 6/20 (30\%) versus 0/10 (0\%) for stack instructions. These are descriptive comparisons of different instruction sets; neither the pooled contrast nor the category breakdown controls for task identity or difficulty.

In a rollout for held-out instruction~4, the policy attempted to pick the blue cube instead of the requested red cube. This is qualitative evidence of a grounding error, not a quantified Intent Score.

Under a pooled independent-binomial approximation, the 95\% Wilson intervals for 20/40 and 6/40 are [35.2, 64.8]\% and [7.1, 29.1]\%, respectively. These approximate intervals do not account for instruction heterogeneity or dependence between trials and do not measure variation across independently trained policies.

\section{QA-20 details}
\label{qa-20 details}

This section provides the complete set of visual question-answering (VQA)
questions referenced in the main text, listed in Table~\ref{tab:scene1_vqa_questions}.

\begin{small}
\setlength{\LTcapwidth}{\textwidth}
\begin{longtable}{@{}c p{0.88\textwidth}@{}}
\caption{Scene 2 visual question-answering questions.}
\label{tab:scene1_vqa_questions} \\
\toprule
\rowcolor{headerblue}
\textbf{ID} & \textbf{Questions} \\
\midrule
\endfirsthead
\toprule
\rowcolor{headerblue}
\textbf{ID} & \textbf{Questions} \\
\midrule
\endhead
\midrule
\multicolumn{2}{r}{\textit{Continued on next page}} \\
\endfoot
\bottomrule
\endlastfoot
\scenerow{Default scene}
1 & What is the object behind the blue cube? \\
\rowcolor{lightgray}
2 & Output the color of the object to the right of the green bowl. \\
3 & Output the color of the leftmost bowl. \\
\rowcolor{lightgray}
4 & Output the color of the rightmost cube. \\
5 & Output the color of the cylinder to the left of the red cube. \\
\rowcolor{lightgray}
6 & Output the color of the object in front of the red cube. \\
7 & Describe the object to the right of the green bowl, such as what it is and what color it is. \\
\rowcolor{lightgray}
8 & What are the red objects in the scene, and what are their positions? \\
9 & Which object is located on the right side of the blue cube? \\
\rowcolor{lightgray}
10 & Output the color of the rightmost cylinder. \\
11 & Looking at the two red objects on the table, which one is on the left and which one is on the right? \\
\rowcolor{lightgray}
12 & Output the color of the object on the right side of the red cylinder. \\
13 & Output the color of the leftmost cylinder. \\
\rowcolor{lightgray}
14 & Where are the two bowls located relative to each other? Which bowl is on the left side? \\
15 & What is the object behind the blue cylinder? \\
\rowcolor{lightgray}
16 & Output the color of the object in front of the red cylinder. \\
17 & Output the color of the object to the right of the red cylinder. \\
\rowcolor{lightgray}
18 & Output the color of the object to the left of the blue cylinder. \\
19 & Output the color of the object to the left of the yellow bowl. \\
\rowcolor{lightgray}
20 & Output the color of the object to the right of the blue cylinder. \\
\end{longtable}
\end{small}

\section{Failure Cases}
\label{failure cases}

\begin{figure}[H]
    \centering
    \includegraphics[width=1.0\linewidth]{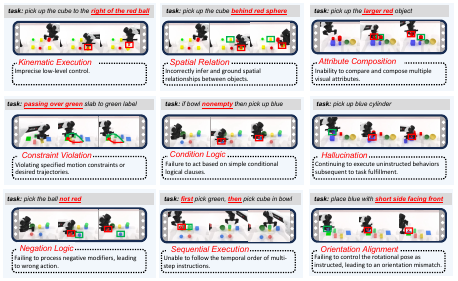}
    \caption{Representative failure cases of the policy. }
    
    \label{fig:failure_cases}
    \vspace{-2em}
\end{figure}

As shown in Fig.~\ref{fig:failure_cases}, the failures fall into two categories. The first occurs when the model understands the task intent but fails due to insufficient execution precision, and the second stems from weaknesses in semantic reasoning and instruction grounding, such as misunderstanding spatial relationships, failing to compare or compose object attributes, inability to meet fine-grained trajectory or orientation constraints, and difficulty handling logical constructs like conditionals, temporal order, and negation. Additionally, the policy may fail to terminate after completing the task, continuing unintended actions not specified in the instruction.

\end{document}